\documentclass[11pt]{article}

\usepackage[preprint]{acl}

\usepackage{times}
\usepackage{latexsym}
\usepackage{forest}
\usepackage{expex}
\usepackage{pgfplots}
\pgfplotsset{compat=1.18}

\usepackage{fvextra}
\DefineVerbatimEnvironment{prompt}{Verbatim}{
  breaklines=true,
  breakanywhere=true,
  fontsize=\small
}

\lingset{
  aboveexskip=0.6ex,
  belowexskip=0.6ex,
  interpartskip=0.2ex,
  glstyle=nlevel,
  glhangstyle=none,
  everygla=\itshape,  
  everyglb=\upshape,  
}

\usepackage[T1]{fontenc}

\usepackage[utf8]{inputenc}

\usepackage{microtype}

\usepackage{inconsolata}

\usepackage{graphicx}

\title{From Lemmas to Dependencies: What Signals Drive Light Verbs Classification?}

\author{
  Sercan Karaka\c{s} \\
  University of Chicago \\
  \texttt{skarakas@uchicago.edu}
  \And
  Yusuf \c{S}im\c{s}ek \\
  F{\i}rat University \\
  \texttt{ysimsek@firat.edu.tr}
}

\begin{document}
\maketitle
\begin{abstract}
Light verb constructions (LVCs) are a challenging class of verbal multiword expressions, especially in Turkish,
where rich morphology and productive complex predicates create minimal contrasts between idiomatic predicate
meanings and literal verb--argument uses. This paper asks what signals drive LVC classification by
systematically restricting model inputs. Using UD-derived supervision, we compare lemma-driven baselines
(lemma TF--IDF + Logistic Regression; BERTurk trained on lemma sequences), a grammar-only Logistic Regression
over UD morphosyntax (UPOS/DEPREL/MORPH), and a full-input BERTurk baseline. We evaluate on a controlled
diagnostic set with Random negatives, lexical controls (NLVC), and LVC positives, reporting split-wise
performance to expose decision-boundary behavior. Results show that coarse morphosyntax alone is insufficient
for robust LVC detection under controlled contrasts, while lexical identity supports LVC judgments but is
sensitive to calibration and normalization choices. Overall, Our findings motivate targeted evaluation of Turkish MWEs and show that ``lemma-only'' is not a single, well-defined representation, but one that depends critically on how normalization is operationalized. We will release the code and dataset to support reproducibility.
\end{abstract}

\section{Introduction}
Multiword expressions (MWEs), which can be described as word sequences that function as partially or fully lexicalized units, are a central property of human language and a persistent challenge for natural language processing, precisely because their meaning often departs from straightforward compositionality and must be learned as a unit \citep{sag2002pain,odijk2013mwe,Ramisch2015,constant-etal-2017-survey,mititelu-etal-2025-survey-mwe}. Recent work further shows that even strong contextualized models struggle to robustly capture idiomatic behavior: transformer-based models treat idiomatic expressions differently from simple lexemes, yet remain insensitive to graded differences in idiomaticity \citep{liu-lareau-2024-assessing}, and VMWE identification continues to require specialized annotation schemes and evaluation pipelines even in high-quality UD resources \citep{markantonatou-etal-2025-vmwe}.

In comprehension and production, speakers are sensitive to distributional regularities of multiword sequences:
frequent or strongly associated expressions can yield measurable processing advantages and facilitate learning,
both in first and second language contexts \citep{ArnonClark2011,SiyanovaChanturiaConklinVanHeuven2011, CastroviejoEtAl2023,SiyanovaChanturiaSonbul2025}.
These behavioral findings motivate a computational perspective in which MWEs are not peripheral exceptions but
core targets for models intended to approximate, support, or evaluate language use. Yet, MWEs systematically
strain surface-based generalization: they can be semantically idiosyncratic, syntactically flexible, and
highly variable under inflection, derivation, and word order alternations \citep{sag2002pain,baldwinkim2010mwe,constant-etal-2017-survey}.

In NLP, MWE processing has traditionally split into \emph{discovery} (mining candidate expressions from corpora)
and \emph{identification} (detecting and typing MWEs in context), with the latter typically formalized as
supervised tagging or structured prediction \citep{Ramisch2015,constant-etal-2017-survey}.
Community benchmarks such as the PARSEME shared tasks have sharpened the field’s focus on \emph{verbal} MWEs
(VMWEs), explicitly targeting discontinuity and morphosyntactic variability across languages
\citep{savary-etal-2017-parseme,savary-etal-2023-parseme13}.
While transformer encoders have become strong default architectures for MWE identification, recent analyses
emphasize that high token-level scores can mask brittle reliance on lexical templates or shallow heuristics,
particularly for idiomatic and semi-lexicalized expressions \citep{premasiri-ranasinghe-2022-berts,miletic-walde-2024-semantics}.
This gap between aggregate accuracy and robust semantic handling is especially salient for evaluation. Since
many surface forms can realize the same content, automatic metrics are often hard to interpret and
human-centered criteria are crucial for defining what constitutes model success \citep{zhou-etal-2022-deconstructing}.

Turkish provides a demanding testbed for these issues. Its rich inflectional and derivational morphology
multiplies surface variation, and productive complex predicate formation yields large families of verb--nominal
predicates that range from transparent compositional combinations to strongly lexicalized idioms
\citep{Oflazer1994,oflazer-etal-2004-integrating,Butt2010LightVerbJungle,Ucar2010LightVerbConstructions}.
Among these, \emph{light verb constructions} (LVCs) are particularly challenging: a nominal element contributes
core predicational content while the verb is partially semantically bleached, and the entire expression behaves
like a single predicate unit \citep{grimshaw1988lightverbs,sag2015complex}.
Crucially, many Turkish light verbs also occur in fully literal transitive or ditransitive uses, creating hard
minimal contrasts in which the same surface frame can express either a conventionalized predicate meaning or a
literal event of transfer, perception, or action. This makes Turkish LVC detection a targeted probe of whether
models encode the distinction between \emph{predicate-level meaning} and \emph{surface argument structure}.
Although Turkish treebanks and MWE resources (including PARSEME) support supervised modeling, annotation
conventions are not always exhaustive and practical performance can depend strongly on preprocessing quality
(e.g., lemma and POS accuracy) \citep{eryigit-etal-2011-multiword,eryigit-etal-2015-annotation,ozturk-etal-2022-enhancing}.

\paragraph{Lexical vs.\ morphosyntactic evidence for LVCs.}
A central open issue in MWE modeling is where the evidence for MWE status resides: in lexical identity
and selection (often well approximated by lemmas), or in morphosyntactic configuration (POS/morphology and
dependency structure). This question mirrors a long-standing debate in psycholinguistics and acquisition
about how multiword knowledge is represented: many accounts argue that speakers acquire and exploit
\emph{stored multiword units} (or strongly entrenched chunks) whose availability is driven by frequency and
association strength, yielding robust processing advantages for familiar sequences
\citep{wray2002formulaic,bannard-matthews2008-stored,arnon-snider2010-morethanwords}.
From this perspective, lemma-based representations are not merely an engineering simplification but a
computational proxy for a lexicalist hypothesis: that MWE/LVC decisions can often be predicted from the
identity of the participating lexemes and their conventionalized co-occurrence patterns.
At the same time, MWEs, and especially verbal MWEs, also exhibit discontinuity, syntactic flexibility, and
morphological variation, which motivates the use of morphosyntactic and dependency information both for
candidate generation and for disambiguation in context
\citep{savary-etal-2017-parseme,simko-etal-2017-uszeged,constant-etal-2017-survey}.
Light verb constructions provide a particularly stringent test of the lexical--morphosyntactic division:
because LVCs often share surface argument-structure frames with fully literal verb--argument configurations,
broad structural cues can be insufficient unless a model also tracks predicate-level meaning and
lexicalized selection restrictions. Consistently, prior work on LVC identification suggests that some cases
are recoverable from syntactic regularities, while others require information beyond syntax alone
\citep{cordeiro-candito-2019-syntax}. Hence, reliable
lemmatization and morphosyntactic preprocessing are themselves non-trivial, and resource-quality analyses for
Turkish VMWE corpora emphasize that lemma/POS accuracy can be a major driver of identification performance
\citep{ozturk-etal-2022-enhancing}. Crucially, recent work suggests that the same linguistic complexity also makes \emph{tokenization} a first-order modeling choice rather than a neutral preprocessing step \citep{bayram2025tokenizationstandards-siu, bayram-etal-2025-tokens-with-meaning}

Moreover, experimental work on Turkish speech formulas and
multiword expressions investigates whether familiar multiword strings show processing advantages over novel
controls \citep{goymen-aygunes2020-speechformulas}.
Meanwhile, Turkish LVCs are highly productive and overlap structurally with literal transitive/ditransitive
uses of the same light verbs, yielding hard minimal contrasts where the same syntactic frame can correspond to
distinct predicate meanings \citep{grimshaw1988lightverbs,Butt2010LightVerbJungle,Ucar2010LightVerbConstructions}.
These properties make Turkish LVC detection an ideal setting for directly probing the relative contributions
of lemma-level lexical content and morphosyntactic structure, including how comprehenders map surface form to
thematic/event interpretation \citep{ozge-unal-bayirli-2022-lightverbs}.

\paragraph{Paper focus.}
Motivated by this, the present paper treats the lexical vs.\ morphosyntactic division not as a secondary
engineering choice but as a primary scientific question for Turkish MWE modeling. We operationalize this
question via controlled literal--idiomatic contrasts and by positioning models along a spectrum of available
information, from lemma-restricted representations (emphasizing lexical selection and chunk-like knowledge)
to morphosyntax-restricted representations (emphasizing grammatical configuration), with the goal of
characterizing which signals are necessary and which are insufficient for robust LVC judgments. More broadly, this framing connects Turkish MWE identification to ongoing work on what kinds of
linguistic structure models encode and how reliably such structure can be extracted
or diagnosed with probing-style methods \citep{hewitt-manning-2019-structural,rogers-etal-2020-primer,
katinskaia-yangarber-2024-probing,yoshida-etal-2024-tree,temesgen-etal-2025-extracting,agarwal-etal-2025-mechanisms,
diego-simon-etal-2025-probing, acevedo2026differential}. In parallel, recent evaluation frameworks emphasize moving beyond
single pooled scores toward more structured and contamination-resistant assessments that better
separate genuine capability from memorization or artifacts \citep{cao-etal-2024-structeval}. 

Our diagnostic evaluation yields three main findings. First, coarse morphosyntactic information alone is largely sufficient to recognize broad negatives but is not sufficient for reliably identifying LVC positives under controlled literal--idiomatic contrasts. Second, lemma-level lexical identity provides substantially stronger evidence for LVC status, but lemma-based classifiers exhibit a calibration trade-off: improving performance on lexical controls can come at the cost of sharply increased false negatives on LVCs. Third, even for contextual encoders trained on lemma sequences, test-time behavior depends on how lemma normalization is instantiated, indicating that ``lemma-only'' is best understood as a family of representations with meaningful distribution shifts rather than a single input setting.

The remainder of the paper is organized as follows. Section~2 presents related work on MWEs/VMWEs and Turkish LVCs, with emphasis on lexical vs.\ morphosyntactic evidence and evaluation practices. Section~3 describes our UD-derived supervision, restricted-input representations, and the construction of the controlled diagnostic set. Section~4 introduces the models and training setups. Section~5 details the experimental protocol and split-wise evaluation. Section~6 reports the results and discusses what they imply about the signals exploited by each model.

\section{Related Work}

As we have shown in the previous section, multiword expressions are not only a modeling nuisance but also a cognitive reality since a substantial portion
of everyday language is produced and understood in partially prefabricated or strongly associated sequences \citep{wray2002formulaic, conklin-schmitt-2012-processing}.
Psycholinguistic and acquisition research has shown reliable processing advantages for frequent multiword units
and sensitivity to phrase-level statistics, consistent with the view that speakers track distributional
regularities beyond single words \citep{wray2002formulaic}.
This line of work motivates treating MWEs as \emph{core} targets for NLP systems that aim to approximate or
support human language use, rather than as peripheral exceptions to compositional grammar.

In NLP, MWE processing has traditionally separated \emph{discovery} (mining candidate expressions via
association measures or pattern heuristics) from \emph{identification} (detecting and typing MWEs in context),
with the latter typically formalized as supervised tagging or structured prediction \citep{Ramisch2015,constant-etal-2017-survey}.
A persistent methodological question concerns \emph{where} the evidence for MWE status resides: in lexical
identity/selection (often approximated by lemmas and collocational strength) versus morphosyntactic
configuration (POS/morphology and dependency structure). Practical systems therefore often combine lexical cues
with morphosyntax and parsing, both to constrain candidate generation and to resolve discontinuity and
variability in context \citep{constant-etal-2017-survey,savary-etal-2017-parseme,simko-etal-2017-uszeged}.

Recent work has largely converged on neural identification architectures, transformer encoders in particular, often benefiting from subword tokenization for robustness to inflectional variation \citep{bui-savary-2024-cross,ide-etal-2025-coam,hefetz-etal-2025-just,rossini-van-der-plas-2026-binary}.
At the same time, several
recent analyses and surveys emphasize that strong aggregate scores can mask brittle reliance on shallow
templates, memorized lexical cues, or dataset-specific artifacts, especially for idiomatic and semi-lexicalized
expressions where the core generalization is semantic \citep{premasiri-ranasinghe-2022-berts,miletic-walde-2024-semantics}.
This concern aligns with a broader trend in contemporary NLP and AI evaluation: controlled test suites and
behavioral-style probes are increasingly used to localize \emph{which} signals a model exploits, rather than
only \emph{how often} it is correct on pooled accuracy. 

\section{Methods}

\subsection{Training supervision from UD treebanks}
To train supervised classification heads, we derived proxy supervision from nine Turkish Universal Dependencies (UD)
treebanks released in CoNLL-U format: UD Turkish-GB, UD Turkish-IMST, UD Turkish-FrameNet, UD Turkish-BOUN,
UD Turkish-PUD, UD Turkish-Penn, UD Turkish-Tourism, UD Turkish-Kenet, and UD Turkish-Atis
\citep{ud-tr-atis,ud-tr-boun,ud-tr-framenet,ud-tr-gb,ud-tr-imst,ud-tr-kenet,ud-tr-penn,ud-tr-pud,ud-tr-tourism}.

These resources, comprising a total of 82,884 sentences, provide UD-standard tokenization, lemmas, UPOS tags, morphological features, and dependency
relations, which we use both to derive weak supervision and to construct restricted-input representations. 

To identify candidate LVC realizations from UD annotations, we exploited the dependency relations
\texttt{compound:lvc} and \texttt{compound}. For treebanks that explicitly
annotate LVCs with \texttt{compound:lvc}, we used these arcs directly as LVC candidates. For treebanks that do
not contain \texttt{compound:lvc}, we followed an alternative procedure: (i) extract all \texttt{compound}
dependencies, (ii) retain only noun--verb \texttt{compound} configurations (a nominal dependent linked to a
verbal head), and (iii) manually review these candidates to remove non-LVC cases and to finalize a
linguistically coherent set of target LVC patterns. Sentences containing validated LVC patterns were labeled
as positive (\texttt{[1]}), and sentences without such patterns were labeled as negative (\texttt{[0]}).

To be even more precise, this pre-filtering step yielded 10{,}056 candidate LVC sentences.
All candidates were then manually verified by two annotators using a binary label (LVC present vs.\ absent).
During this screening, we identified 565 candidates that were incorrectly labeled as LVCs and removed them from the dataset.
The resulting dataset contains 82{,}319 sentences, of which 9{,}491 contain at least one LVC.

\subsection{Restricted-input representations}
It is worth recalling that a central methodological choice is to \emph{restrict} what information the model can access in order to probe
the relative contributions of lexical identity versus coarse morphosyntactic
structure.

For lemma-restricted baselines, each sentence is converted to a sequence of lemma strings (preserving token
order). This removes surface wordform variation while retaining lexical identity in normalized form and
allowing contextual models to compose over the lemma sequence.

For grammar-restricted baselines, we remove lexical identity and keep only UD-style grammatical information:
UPOS tags, dependency relations (\texttt{DEPREL}), and morphological features (\texttt{FEATS}). Each sentence
is represented as a sentence-level count vector over these features (a bag of UPOS/DEPREL/MORPH features).
Training uses the UD-derived supervision described above. At evaluation time, UD-style annotations for the
diagnostic set are produced automatically, using Stanza \citep{qi-etal-2020-stanza}, and features are extracted from these predicted
annotations.

\subsection{Task and evaluation set}
Recall that we frame Turkish light verb construction detection as binary sentence classification:
\texttt{[1]} indicates an LVC (including lexicalized/idiomatic noun+verb predicates under our labeling policy),
and \texttt{[0]} indicates a literal verb--argument configuration. The controlled diagnostic evaluation set contains three balanced conditions ($n{=}49$ each; $N{=}147$):
(i) \textsc{Random} negatives: sentences that contain no LVC and are not constructed to reuse the specific
light-verb lemmas targeted by the positive set; (ii) \textsc{NLVC} (\emph{non-LVC}) lexical controls:
in-domain sentences that reuse the same light-verb lemmas as the positives but enforce a \emph{literal
(non-idiomatic)} verb--argument reading; and (iii) \textsc{LVC} positives: sentences that contain an intended idiomatic
light verb construction.

All evaluation items were manually written and then validated by three independent annotators for linguistic
naturalness, pragmatic plausibility, and label agreement. This design prioritizes minimizing annotation noise
and avoiding uncontrolled ambiguity over maximizing corpus breadth or domain diversity. Importantly, the diagnostic set was created specifically for this study rather than sampled from an existing
treebank or previously annotated corpus. 

Although LVCs are often modeled via token-level identification (e.g., sequence tagging or arc labeling), we
formulate the task as sentence-level \emph{presence} detection to keep the comparison across restricted-input
representations clean and assumption-light (no span,
discontinuity, or decoding choices). This is especially relevant in Turkish because the same light verb can
occur both in conventionalized LVC-like predicates and in fully literal uses (e.g., \textit{vermek} `give' in
\textit{öpücük vermek} `give a kiss' vs.\ literal transfer uses), creating hard minimal contrasts for Turkish \citep{ ozge-unal-bayirli-2022-lightverbs} and other languages \citep{wittenberg-pinango-2011-lvc, wittenberg-etal-2014-rose-kiss, wittenberg-snedeker-2014-kiss}.

Despite the fact that $N=147$ may appear modest, the dataset is purpose-built as a \emph{controlled diagnostic} rather than an i.i.d.\ benchmark: it is balanced across conditions and lexically matched to isolate the LVC vs.\ literal distinction while minimizing confounds from topic, vocabulary, or annotation noise. This design choice follows a growing evaluation philosophy that prioritizes targeted stress tests and behavioral suites over aggregate i.i.d.\ accuracy, including checklist- and template-driven behavioral testing \citep{ribeiro-etal-2020-beyond,naik-etal-2018-stress,yang-etal-2022-testaug, zhao-etal-2024-syntheval}, contrast sets that probe local decision boundaries \citep{gardner-etal-2020-evaluating}, and dynamic/challenge-style benchmarks that surface systematic failure modes \citep{kiela-etal-2021-dynabench}. Related lines of work similarly argue that standard test splits can mask reliance on brittle heuristics or shortcuts \citep{mccoy-etal-2019-right,linzen-2020-accelerate,geirhos-etal-2020-shortcut}. Finally, minimal-pair and diagnostic evaluation traditions show that carefully curated item sets can be disproportionately informative for pinpointing specific linguistic generalizations \citep{marvin-linzen-2018-targeted,warstadt-etal-2020-blimp-benchmark,mueller-etal-2020-cross, mayne-etal-2025-decision-boundaries, karakas2026}.

\section{Models}

\subsection{Supervised BERTurk (full-input baseline)}
We fine-tune BERTurk \texttt{32K cased} and \texttt{128K cased} \citep{dbmdz-berturk} by adding a task-specific binary
classification head over the final-layer \texttt{[CLS]} representation. Training uses UD-derived proxy
supervision with an 80/20 train/test split under stratified sampling to preserve the class distribution.
For tokenization, we apply standard padding and truncation with a maximum sequence length of 64 tokens.
To reduce overfitting, we set both hidden dropout and attention dropout to 0.2. Models are trained with
learning rate $2\times10^{-5}$, batch size 32, and weight decay 0.01 for up to 10 epochs, monitoring
validation loss with early stopping (patience 3) and selecting the checkpoint with the lowest validation
loss. We report Accuracy, F1, and Loss. BERTurk 32K achieves 93.94\% Accuracy, 0.7558 F1, and 0.1500 Loss;
BERTurk 128K achieves 94.06\% Accuracy, 0.7508 F1, and 0.1506 Loss, indicating comparable performance across
vocabulary sizes. Notably, this baseline uses the original sentence forms as input (no lemmatization).

\subsection{Lemma-only Logistic Regression (TF--IDF)}
As a lightweight lexical baseline, we train Logistic Regression over TF--IDF features computed from lemma
$n$-grams. Each sentence is replaced by its lemma sequence, and the resulting lemma text is lowercased
prior to vectorization. We use unigram+bigram lemma features (\texttt{ngram\_range=(1,2)}) with
\texttt{max\_features=5000}. The classifier is $L_2$-regularized Logistic Regression trained with
\texttt{class\_weight="balanced"} to mitigate class imbalance; we set \texttt{max\_iter=1000} to ensure
stable convergence. Supervised training follows the same 80/20 stratified split procedure. Evaluation is
reported via overall accuracy and a standard classification report; model predictions for the diagnostic
test set are exported for analysis.

\subsection{Lemma-only BERTurk (restricted lexical input)}
To isolate the effect of contextual composition under lemma normalization, we fine-tune BERTurk classifiers
on inputs where each sentence is replaced by its lemma sequence, preserving token order and
context while removing surface wordforms. This setting retains sequential context while explicitly removing
surface-form variation due to inflection and orthography. We report lemma-only results for both BERTurk
variants (32K and 128K) under the same training protocol as the full-input baseline (stratified 80/20 split,
max length 64, padding/truncation, dropout 0.2, learning rate $2\times10^{-5}$, batch size 32, weight decay
0.01, early stopping with patience 3).

\subsection{Grammar-only Logistic Regression (UPOS/DEPREL/MORPH)}
To probe how far coarse morphosyntactic information goes without lexical identity, we train a Logistic
Regression model using only UD-style grammatical annotations: UPOS tags, dependency relations, and
morphological features. From the UD treebank training data, we extract (i) each UPOS tag, (ii) each dependency
relation label, and (iii) each morphological key--value pair as a separate feature, and we construct a fixed
feature inventory \emph{only from the training portion} of the supervision. Each sentence is then represented
as a sentence-level frequency vector counting occurrences of these UPOS/DEPREL/MORPH
features. The classifier is trained as a binary Logistic Regression model with \texttt{class\_weight="balanced"}
and \texttt{max\_iter=2000}.

\section{Experiments}
\label{sec:experiments}

Our experiments aim to identify which information sources underlie Turkish LVC judgments by systematically restricting model inputs. We compare models along a continuum from lexical, lemma-based representations (lemma-only LR; lemma-only BERTurk) to grammar-only representations (LR over UPOS/DEPREL/MORPH), and include a full-input contextual baseline (BERTurk on surface sentences). This design enables a controlled assessment of normalized lexical identity, coarse morphosyntactic cues without lexical content and unrestricted contextual encoding.

\subsection{Diagnostic evaluation set and split-wise reporting}
We evaluate on a purpose-built diagnostic set of $N{=}147$ sentences, balanced across three conditions ($n{=}49$ each): \textsc{Random} negatives, \textsc{NLVC} lexical controls that reuse the same light-verb lemmas in literal predicate--argument contexts, and \textsc{LVC} positives. This setup targets the key contrast---conventionalized predicate meaning vs.\ literal uses while controlling salient lexical cues via \textsc{NLVC}. Because we probe decision boundaries under minimal contrasts rather than maximize i.i.d.\ coverage, we report split-wise (Random/NLVC/LVC) results in addition to pooled scores; pooled accuracy can mask systematic misses of \textsc{LVC} under a conservative negative bias.

For the full-input BERTurk baseline, we evaluate each item in its original surface form (inflected wordforms), mirroring standard VMWE/LVC identification.

Lemma-only BERTurk models are trained on lemmatized sentences (lemma sequences) and evaluated under two conditions: (i) \emph{surface-test}, testing on original surface sentences (intentional train--test mismatch), and (ii) \emph{lemma-test}, testing on a lemmatized version of the diagnostic set. This probes whether lemma-only models learn a stable lemma-level decision function or are sensitive to distribution shift at test time.

Grammar-only LR is trained on sentence-level count vectors over \textsc{UPOS}/\textsc{DEPREL}/\textsc{MORPH} features from UD training data. Since the diagnostic items are not from UD, we obtain UD-style annotations with Stanza and featurize them using the fixed training-only feature inventory; we additionally manually checked Stanza outputs for all 147 items to guard against systematic tagging/feature errors.

\section{Results and Discussion}
\label{sec:results}

\begin{table*}[t]
  \centering
  \small
  \setlength{\tabcolsep}{3.5pt}
  \renewcommand{\arraystretch}{1.08}
  \begin{tabular}{llccccrrcc}
    \hline
    Model & Setting & Random & NLVC & LVC & Overall & FP & FN & Prec & Rec \\
    \hline

    \multicolumn{10}{l}{Lemma-only LR } \\
    \hspace{1em} & Run 1 & 98.0 & 75.5 & 49.0 & 74.1 & 13 & 25 & 64.9 & 49.0 \\
    \hspace{1em} & Run 2 & 98.0 & 87.8 & 32.7 & 72.8 &  7 & 33 & 69.6 & 32.7 \\
    \hspace{1em} & Run 3 & 100.0 & 95.9 & 18.4 & 71.4 &  2 & 40 & 81.8 & 18.4 \\
    \hline

       \multicolumn{10}{l}{BERTurk Full} \\
    \hspace{1em}32k   & Surface test & 98.0  & 81.6 & 67.3 & 82.3 & 10 & 16 & 76.7 & 67.3 \\
    \hspace{1em}128k  & Surface test & 98.0  & 81.6 & 79.6 & 86.4 & 10 & 10 & 79.6 & 79.6 \\
    \hline
    \hspace{1em}32k   & Lemma test   & 100.0 & 79.6 & 63.3 & 81.0 & 10 & 18 & 75.6 & 63.3 \\
    \hspace{1em}128k  & Lemma test   & 100.0 & 81.6 & 61.2 & 81.0 &  9 & 19 & 76.9 & 61.2 \\
    \hline

    \multicolumn{10}{l}{BERTurk (lemma-only)} \\
    \hspace{1em}32K  & Surface-test & 98.0 & 87.8 & 67.3 & 84.4 &  7 & 16 & 82.5 & 67.3 \\
    \hspace{1em}128K & Surface-test & 98.0 & 83.7 & 71.4 & 84.4 &  9 & 14 & 79.5 & 71.4 \\
    \hline
    \hspace{1em}32K  & Lemma-test   & 100.0 & 91.8 & 14.3 & 68.7 &  4 & 42 & 63.6 & 14.3 \\
    \hspace{1em}128K & Lemma-test   & 100.0 & 87.8 & 36.7 & 74.8 &  6 & 31 & 75.0 & 36.7 \\
    \hline

    \multicolumn{10}{l}{Grammar-only LR} \\
    \hspace{1em} & -- & 100.0 & 91.8 & 10.2 & 67.3 &  4 & 44 & 55.6 & 10.2 \\
    \hline
  \end{tabular}
  \caption{\label{tab:diag-restricted}
    Restricted-input supervised baselines on the diagnostic set.
    Entries are success rates (\%) for each condition ($n=49$) and Overall ($N=147$).
    FP/FN are pooled error counts where FP counts mistakes on negatives (Random+NLVC) and FN counts mistakes on LVC.
    Precision/Recall treat LVC as the positive class and pool Random+NLVC as negatives.
  }
\end{table*}

Lemma-only LR exhibits a clear calibration trade-off across runs: Runs~1--3 reflect different decision
thresholds on the model’s posterior $p(y{=}1\mid x)$ (Run~1 uses the default $\tau{=}0.5$, while Runs~2--3 use
increasingly conservative thresholds $\tau^\star$ chosen by sweeping $\tau\in[0,1]$ on a held-out split using
the precision--recall curve—e.g., selecting the $\tau^\star$ that maximizes LVC $F_1$ and a higher-$\tau$
setting that prioritizes precision). As the classifier becomes more conservative, accuracy on negatives
improves. Random remains near ceiling and NLVC can be pushed high while LVC accuracy drops markedly due to a
rise in false negatives (Table~\ref{tab:diag-restricted}). This pattern is consistent with a decision
boundary that defaults to \texttt{[0]} when lexical evidence is weak, yielding strong performance on negatives
but limited sensitivity to LVC positives.

The grammar-only LR baseline makes the limitation of coarse morphosyntax especially explicit: despite
near-perfect performance on Random and strong performance on NLVC, it collapses on LVC positives
(Table~\ref{tab:diag-restricted}). Given that (i) the diagnostic items are constructed to create minimal
literal--idiomatic contrasts and (ii) Stanza analyses were manually checked for all 147 items, the most
straightforward interpretation is that UPOS/DEPREL/MORPH frequency signatures alone are not sufficient to
recover LVC status in these controlled contrasts. The model can learn broad grammatical regularities that
support negative decisions, but it largely fails to identify predicate-level conventionalization without
lexical identity.

\subsection{Lemma-only BERTurk: test-time input form matters}
Lemma-only BERTurk shows a strong dependence on test-time input form (Table~\ref{tab:diag-restricted}). When
evaluated on surface sentences (surface-test), lemma-only BERTurk attains high accuracies across splits,
including comparatively strong LVC performance. This suggests that even when supervision is provided on lemma
sequences, the fine-tuned encoder can often generalize to inflected surface forms at inference time---a
plausible outcome given subword tokenization overlap between lemmas and inflected variants.

However, when the same lemma-only models are evaluated on the lemmatized version of the diagnostic set
(lemma-test), LVC accuracy drops sharply (especially for 32K; Table~\ref{tab:diag-restricted}). This
asymmetry suggests distribution shift between the lemma representation used for training and the lemma
representation used at diagnostic evaluation. In our setup, lemma-only BERTurk is trained on UD-derived lemmas
(treebank normalization conventions), whereas the lemmatized diagnostic set is produced by an automatic
pipeline. Small mismatches in lemma inventory (e.g., normalization, casing, segmentation choices, named-entity
handling, or token boundary differences) can substantially change the subword sequences seen by the model,
making lemma-test a different input distribution rather than a uniformly easier condition.

The 32K vs.\ 128K contrast is also suggestive: the 128K lemma-only model degrades less under lemma-test than the
32K model (Table~\ref{tab:diag-restricted}; LVC recall 36.7\% vs.\ 14.3\%), consistent with the idea that a larger
subword vocabulary can reduce token fragmentation (i.e., splitting lemmas into many smaller pieces) and thereby
improve robustness to normalization- or lemma-form mismatches
\citep{sennrich-etal-2016-neural, devlin-etal-2019-bert, provilkov-etal-2020-bpe, chizhov-etal-2024-bpe}.
Comparing BERTurk full and lemma-only under the same surface evaluation further reveals a vocabulary-dependent
trade-off (Table~\ref{tab:diag-restricted}): with 32K, lemma-only matches LVC recall while improving performance on
negatives (higher NLVC accuracy and precision), yielding a slightly higher Overall score and fewer false positives.
With 128K, the full model is clearly stronger on LVCs (higher recall, fewer false negatives) and improves Overall,
consistent with surface-form supervision providing useful morpho-lexical cues when subword coverage is better.
Notably, the 32K models show the opposite pattern: under surface-test, lemma-only slightly outperforms full overall
despite identical LVC recall (Table~\ref{tab:diag-restricted}), suggesting that at smaller subword vocabularies,
incorporating surface-form detail can increase spurious matches on negatives without improving sensitivity to LVCs.

Evaluating the full BERTurk models on lemma-normalized inputs further sharpens this interpretation
(Table~\ref{tab:diag-restricted}). Under lemma-test, both full models obtain the same Overall accuracy (81.0\%),
but LVC recall decreases relative to surface-test. Somewhat surprisingly, the 128K full model, despite being the
stronger surface-form model, shows a much larger recall drop under lemmatization than the 32K full model
(Table~\ref{tab:diag-restricted}). This pattern suggests that the 128K model's surface-form gains may rely more
heavily on morpho-lexical cues present in inflected surface strings, which are attenuated or removed by
lemma normalization. At the same time, full-on-lemma-test remains clearly stronger than lemma-only-on-lemma-test in
LVC recall (Table~\ref{tab:diag-restricted}), indicating that restricting training inputs to lemmas does not by
itself yield robustness on LVCs in this diagnostic setting.

To sum up, the diagnostic results support three claims that motivate the paper.
First, pure grammar signals (coarse UPOS/DEPREL/MORPH frequencies) can be sufficient to recognize negatives but
are insufficient for robust LVC detection under controlled literal--idiomatic contrasts. Second, lemma-level
lexical identity supports LVC judgments, but classical bag-of-lemmas models can be highly sensitive to
calibration. Third, even when models are trained on lemma sequences, their behavior depends on how lemma
normalization is instantiated at test time, which can be thought as a piece of evidence that lemma-only is not a single setting but a family
of representations with potentially meaningful distribution shifts.

These patterns connect to several well-established threads in the MWE/VMWE literature.
First, the weak performance of a grammar-only signal on LVC positives is consistent with the shared-task and
survey view that VMWE identification typically requires combining lexical cues with morphosyntactic information:
syntax helps with candidate generation and handling discontinuity/variability, but is often insufficient to
separate idiomatic vs.\ literal readings when they share similar structural frames
\citep{constant-etal-2017-survey,savary-etal-2017-parseme,simko-etal-2017-uszeged,cordeiro-candito-2019-syntax}.
Second, the strong dependence of lemma-only behavior on the exact normalization pipeline aligns with work
highlighting that VMWE performance can be heavily shaped by preprocessing quality and annotation/representation
choices, including Turkish-focused analyses that explicitly examine corpus/preprocessing effects for Turkish
VMWEs \citep{ozturk-etal-2022-enhancing}.
Finally, the divergence between surface-test and lemma-test behavior reinforces a broader evaluation lesson
in contemporary NLP: aggregate scores can obscure what cues models actually exploit, motivating controlled
contrast sets and probing-style analyses that target decision boundaries and representation content
\citep{hewitt-manning-2019-structural, gardner-etal-2020-evaluating,rogers-etal-2020-primer}.

\section{Conclusion}

This paper probed which signals support Turkish light verb construction (LVC) detection under controlled
literal--idiomatic contrasts by restricting model inputs to lemmas, to grammar-only morphosyntax, or to full
surface sentences. The results indicate that coarse UPOS/DEPREL/MORPH summaries can support negative decisions
but are not sufficient for reliably recognizing LVC positives in this diagnostic setting, whereas lemma-level
lexical information is more directly aligned with the predicate-level distinction the task requires.
Results with BERTurk further underscore a vocabulary-dependent trade-off: lemma-only models can reduce false
positives on negatives under surface evaluation, while full surface supervision can substantially improve LVC
sensitivity when subword coverage is strong (Table~\ref{tab:diag-restricted}). However, evaluating full models on
lemma-normalized inputs reveals a non-trivial distribution shift, suggesting that high surface-form performance
can rely on morpho-lexical cues that are attenuated by lemmatization.

At the same time, lemma-based modeling is not monolithic: bag-of-lemmas classifiers are sensitive to
calibration, and lemma-only contextual encoders can be highly sensitive to how lemma normalization is
operationalized at test time, revealing distribution-shift effects. These findings motivate split-wise
reporting (Random/NLVC/LVC) as a default for LVC evaluation and position Turkish LVCs as a useful probe for
separating lexicalized predicate meaning from surface argument structure in morphologically rich languages.

\section*{Limitations}

Our supervision is UD-derived (\texttt{compound:lvc} and noun--verb \texttt{compound}), which is a practical
proxy but may reflect treebank-specific conventions and incomplete coverage. The diagnostic set is small
($N{=}147$) by design, so results should be read as decision-boundary evidence under controlled contrasts, not
as in-the-wild performance. Grammar-only evaluation relies on Stanza parses (manually checked but not error
free). Finally, lemma-only results can be sensitive to lemmatization/tokenization conventions, so “lemma-test”
effects may partly reflect normalization choices rather than model competence alone.

\bibliography{custom}

\clearpage
\appendix

\clearpage
\appendix
\onecolumn
\end{document}